\documentclass[10pt,twocolumn,letterpaper]{article}

\usepackage[pagenumbers]{cvpr} 

\usepackage{gradient-text}

\newcommand{\addcite}[1]{{\textcolor{blue}{[CITE]}}}

\def\CircleArrowright{\ensuremath{%
  \rotatebox[origin=c]{310}{$\circlearrowright$}}}
\newcommand{\vlnbert}{VLN$\protect\CircleArrowright$BERT}

\newcommand{\cmark}{\ding{51}}%
\newcommand{\objnav}{\textsc{ObjectNav}\xspace}

\newcommand{\ours}{SAME}

\definecolor{cvprblue}{rgb}{0.21,0.49,0.74}
\usepackage[pagebackref,breaklinks,colorlinks,allcolors=cvprblue]{hyperref}

\usepackage[dvipsnames]{xcolor}
\usepackage{colortbl}
\usepackage{multirow}
\usepackage{amssymb}
\usepackage{amsmath}
\usepackage{wrapfig}
\usepackage{float}
\usepackage{pifont}

\def\paperID{278} 
\def\confName{CVPR}
\def\confYear{2025}

\title{\textit{\gradientRGB{SAME}{0,0,255}{77, 163, 193}}: Learning Generic Language-Guided Visual Navigation \\ with State-Adaptive Mixture of Experts}

\author{%
  Gengze Zhou$^1$\quad
  Yicong Hong$^2$\quad
  Zun Wang$^3$\quad
  Chongyang Zhao$^4$\quad
  Mohit Bansal$^3$\quad
  Qi Wu$^1$ \\
  $^1$The University of Adelaide\quad
  $^2$Adobe Research \quad
  $^3$UNC, Chapel Hill \quad
  $^4$UNSW Sydney\\
  \tt\small {\{gengze.zhou, qi.wu01\}@adelaide.edu.au}\\
{\small \url{https://github.com/GengzeZhou/SAME}}
}

\begin{document}
\maketitle
\begin{abstract}
The academic field of learning instruction-guided visual navigation can be generally categorized into high-level category-specific search and low-level language-guided navigation, depending on the granularity of language instruction, in which the former emphasizes the exploration process, while the latter concentrates on following detailed textual commands. Despite the differing focuses of these tasks, the underlying requirements of interpreting instructions, comprehending the surroundings, and inferring action decisions remain consistent.
This paper consolidates diverse navigation tasks into a unified and generic framework -- we investigate the core difficulties of sharing general knowledge and exploiting task-specific capabilities in learning navigation and propose a novel State-Adaptive Mixture of Experts (SAME) model that effectively enables an agent to infer decisions based on different-granularity language and dynamic observations. Powered by SAME, we present a versatile agent capable of addressing seven navigation tasks simultaneously that outperforms or achieves highly comparable performance to task-specific agents.
\end{abstract}    
\section{Introduction}
\label{sec:intro}

\begin{figure*}[h]
  \centering
   \includegraphics[width=\linewidth]{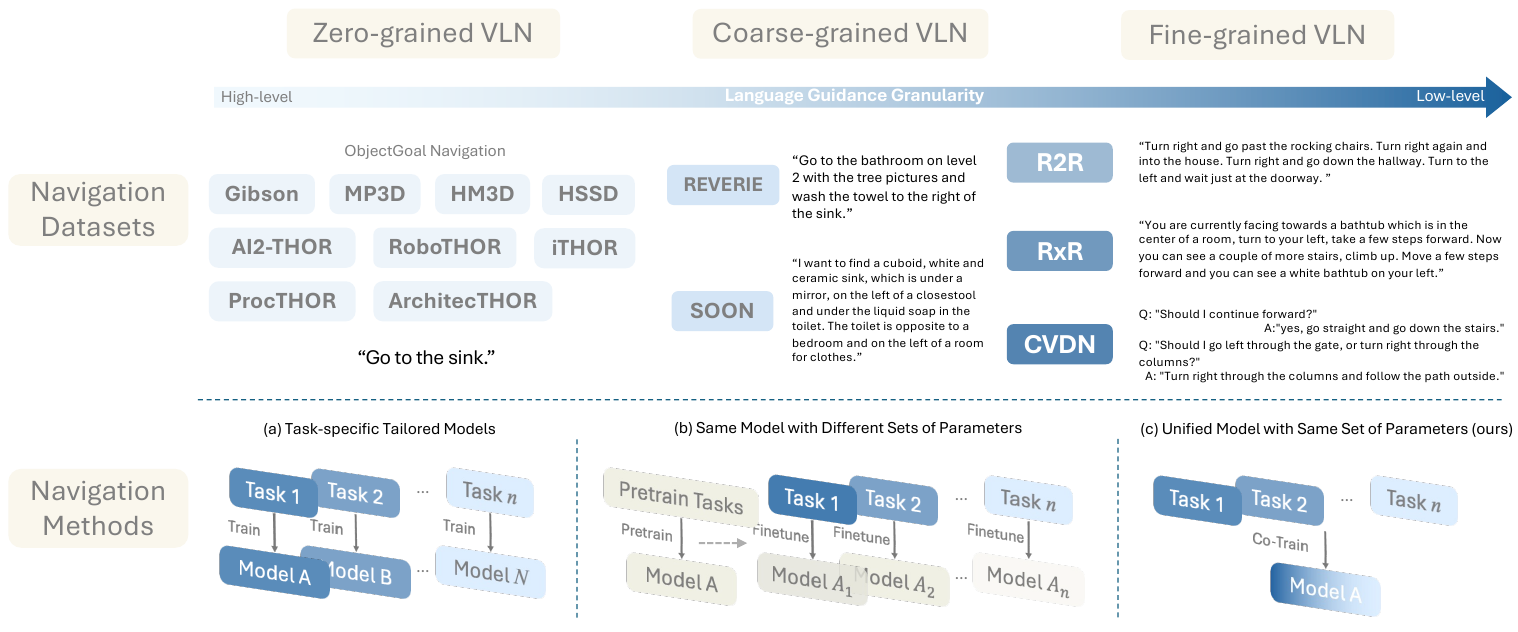}
   \caption{We consolidate diverse navigation tasks into a unified language-guided navigation framework sorted by language granularity. Previous approaches utilize task-specific designs tailored to address particular types of language instructions, as shown in (a) and (b). In contrast, we propose a versatile system that can interpret and execute arbitrary language instructions as shown in (c).}
   \label{fig:teaser}
   \vspace{-5pt}
\end{figure*}

Acquiring the capability to understand natural language commands and navigate in unfamiliar environments constitutes a fundamental competency for embodied intelligence.
In recent years, a great variety of navigational tasks has emerged, each defined by distinct navigation objectives, from broad, high-level goals~\cite{batra2020objectnav, zhu2017target} to detailed, low-level directives~\cite{anderson2018r2r, qi2020reverie, anderson2020rxr, zhu2021soon, thomason2020cvdn}, highlighting exploration and instruction-following, respectively. However, these tasks are mostly formulated as isolated research problems, and the specialized methods developed for each are typically not generalizable to others (Figure~\ref{fig:teaser}a). For example, structured memory tailored for efficient target object exploration~\cite{ramakrishnan2022poni, chaplot2020neural, chaplot2020object}, contextual guidance for vague instructions~\cite{lin2021scene, gao2021room, sigurdsson2023rrex}, and episodic vision-language alignment for instruction following~\cite{fried2018speaker, tan2019envdrop, wang2019reinforced, hu2019you}. Subsequent works leverage generic vision-language representations~\cite{chen2020uniter, li2020oscar, su2019vl, li2019visualbert, tan2019lxmert} to pretrain vision-language-action policies~\cite{hao2020prevalent, hong2020recurrent, guhur2021airbert, majumdar2020improving, chen2021hamt, chen2022duet, qiao2023hop+, li2019robust} (Figure~\ref{fig:teaser}b), finetuning parameters for specific tasks while maintaining the same model architecture.
In this paper, we argue that the essential difference between these tasks lies in the granularity of instruction, and the learning problems should be unified under the broader concept of language-guided visual navigation (VLN), where the overarching goal is to create a versatile system that can interpret and execute arbitrary language instructions (Figure~\ref{fig:teaser}c). 

Apart from the practical perspective, in which the agent's capability should not be constrained by the user's instruction format, unifying navigational problems can potentially create an effective learning paradigm that enables training an agent on a larger amount of data covering wider skill sets. An intuitive approach towards this is to integrate distinct navigational datasets and use them to train an existing agent for multitasking; however, our study reveals that simply mixing data yields inconsistent performance variation across tasks due to conflicting learning objectives (\S\ref{sec:conflicts}). As a result, we resort to a more adaptive framework capable of sharing general knowledge and retrieving task-specific skills to infer appropriate navigation decisions.

Inspired by recent success in the Mixture of Experts (MoE)~\cite{jacobs1991adaptive, jordan1994hierarchical, shazeer2017outrageously} approach for natural language processing, in particular, enhancing the transformer-based Large Language Models~\cite{xue2024openmoe, jiang2024mixtral, dai2024deepseekmoe}, we incorporate MoE to build our generalizable navigation agent.
Unlike previous works that typically implement task-wise or token-wise MoE, which we found ineffective in multi-task navigation learning, we propose a novel MoE formulation for sequential embodied agents in which the experts are selected based on the agent's state (i.e., attended language and visual observation at a certain timestep, see \S\ref{sec:moe_type}). We further show that applying MoE on visual queries gives better results than applying it on feed-forward networks as in LLMs, aligning with the fact that many navigational agents rely on multi-view visual attention for decision-making.

We termed this method \textit{State-Adaptive Mixture of Experts}, or \textit{SAME}, suggesting applying the \textit{same} model to solve a wide range of navigation problems. Powered by SAME, we train a versatile agent across seven major language-guided navigation tasks, including R2R~\cite{anderson2018r2r}, RxR-EN~\cite{ku2020room}, REVERIE~\cite{qi2020reverie}, \objnav\footnote{We consider Object-Goal Navigation~\cite{batra2020objectnav} as a form of zero-grained language-guided navigation, as will be specified in \S\ref{subsec:formulation}. \label{footnote_objnav}}, CVDN~\cite{thomason2020cvdn}, SOON~\cite{zhu2021soon}, and R2R-CE~\cite{krantz2020beyond}, achieving state-of-the-art or highly comparable performance to models tailored for a single task.
\section{Background}
In this section, we begin by introducing the formulation of language-guided visual navigation tasks under varying levels of language granularity. We employ DUET~\cite{chen2022duet} to illustrate a general cross-modality navigation model for language-guided navigation. Through a series of contrastive experiments, we analyze the interconnections among these navigation tasks, understanding their underlying contradictions and providing a proof of concept for our approach.

\subsection{Navigation Tasks Formulation}
\label{subsec:formulation}
Given an instruction represented by a sequence of $L$ word embeddings $\mathcal{W} = \{w_i\}_{i=1}^L$, an agent navigates on a predefined undirected graph $\mathcal{G} = \langle \mathcal{V}, \mathcal{E} \rangle$, where $\mathcal{V}$ represents the navigable nodes and $\mathcal{E}$ represents the connectivity edges. The agent is expected to execute a sequence of actions $ \{s_0, a_0, s_1, a_1, \ldots, s_T, a_T\} $ to navigate to the target position $ v_T $, as specified by the instruction $ \mathcal{W} $. Each action $a_t$ transit the agent from the current state $ s_t = \langle v_t, \theta_t, \phi_t \rangle $ to $ s_{t+1} = \langle v_{t+1}, \theta_{t+1}, \phi_{t+1} \rangle $ which includes its spatial location $ v_t \in \mathcal{V} $, heading angle $ \theta_t $, and elevation angle $ \phi_t $, and generate a new visual observation $\mathcal{O}_t$. Additionally, the agent maintains a record of the state history $ h_t $ and adjusts the conditional transition probability between states, defined as $ \mathcal{S}_t = \mathcal{T}(s_{t+1} \mid a_t, s_t, h_t, \mathcal{O}, \mathcal{W}) $, where $ \mathcal{T} $ denotes the conditional transition probability distribution. We categorize the language instruction $\mathcal{W}$ into three classes by granularity as:
\begin{itemize}
    \item \textbf{Fine-grained VLN}: $\mathcal{W}$ describes the sequence of actions $ \{s_0, a_0, s_1, a_1, \ldots, s_T, a_T\} $ step-by-step.
    \item \textbf{Coarse-grained VLN}: $\mathcal{W}$ refers to an out-of-sight target at $v_T$, \textit{e.g.}, ``the cold tap in the first bedroom on level two".
    \item \textbf{Zero-grained VLN}: $\mathcal{W}$ refers to a single term indicating the target (\textit{e.g.}, an object category in \objnav).
\end{itemize}

\noindent\textbf{Multimodal Navigation Policy}
At each step, the agent receives a local visual observation $ \mathcal{O}_t = \{o_i\}_{i=0}^{36} $, consisting of 36 view images, and a language instruction $ \mathcal{W} $. These are encoded separately by a vision encoder and a language encoder into visual feature $ \hat{\mathcal{O}_t} $ and language feature $ \hat{\mathcal{W}} $. DUET~\cite{chen2022duet} incorporates $\hat{\mathcal{O}_t}$ and agent's state $s_t$ to obtain node embedding $\hat{V_t}$ and maintain a topological map $\hat{\mathcal{G}_t} = \{\hat{V_i}\}_{i\leq t}$ as navigation history, details are provided in the supplementary. A local cross-modal encoder is utilized to excite visual features conditioned on language features:
\begin{equation}
    \text{CrossAttn}(\hat{\mathcal{O}_t}, \hat{\mathcal{W}}) = \text{Softmax}\left( \frac{\hat{\mathcal{O}_t}W_q(\hat{\mathcal{W}}W_k)^T}{\sqrt{d}}\right)\hat{\mathcal{W}}W_v,
\end{equation}
The output embedding from the final layer of view $o_i$ is denoted as $\hat{o}_i'$.
Similarly, a parallel global cross-modal encoder is implemented to encode language-conditioned map $\hat{\mathcal{G}_t}' = \text{CrossAttn}(\hat{\mathcal{G}_t},\hat{\mathcal{W}})$. Denote the output embedding of node $V_i$ as $\hat{v}_i'$, the navigation score given by the local and global cross-modal encoder is calculated as:
\begin{align}
    s_i^l &= \text{FFN}^l(\hat{o}_i'), s_i^g = \text{FFN}^g(\hat{v}_i'), \\
    s_i &= \sigma_ts_i^l + (1-\sigma_t)s_i^g
\end{align}
where FFN is a two-layer feed-forward network and $\sigma$ is a learnable parameter.

\noindent\textbf{Datasets}
We select three typical datasets for our contrastive experiments based on instruction granularity:

\begin{itemize}
    \item R2R~\cite{anderson2018r2r}: The fine-grained VLN task which consists of 22k human-annotated navigational instructions. On average, an instruction contains 32 words, and the ground-truth path is formed by 7 steps, totaling 10 meters.
    \item REVERIE~\cite{qi2020reverie}: The coarse-grained VLN task inherits the trajectories from R2R but provides high-level instructions that describe a target object. On average, instructions contain 21 words, and the length of ground-truth paths ranges from 4 to 7 steps.
    \item \objnav-MP3D~\cite{batra2020objectnav}: We use the standard split of 11 validation scenes from the Habitat \objnav dataset~\cite{savva2019habitat} in MP3D~\cite{chang2017matterport3d}, which consists of 21 goal categories. We utilized human demonstration from Habitat-Web~\cite{ramrakhya2022habitatweb} as training data, details are discussed in Section \ref{sec:conflicts}.
\end{itemize}

\noindent\textbf{Evaluation Metrics}
We follow 5 standard metrics in VLN literature to assess the agent performance, including Trajectory Length (TL), Navigation Error (NE), Success Rate (SR), normalized inverse of the Path Length (SPL)~\cite{anderson2018spl}, normalized Dynamic Time Warping (nDTW)~\cite{ilharco2019general}.

\subsection{What are the Conflicts in Navigation Multi-tasks Learning?}
\label{sec:conflicts}

To successfully train a versatile navigation agent, it is essential to understand the underlying contradictions that prevent unified model learning and to gain insights into the best practices for learning from diverse data sources. Following the discussion in Section \ref{sec:intro}, we unify the training data by transferring 70k human demonstration \objnav data from Habitat-Web into trajectories in the discrete environment.

\begin{table}[t]
\centering

\resizebox{\columnwidth}{!}{
\definecolor{Gray}{gray}{0.94}
\definecolor{textGray}{gray}{0.6}

\begin{tabular}{ccccc>{\columncolor{Gray}}c>{\columncolor{Gray}}ccc>{\columncolor{Gray}}c>{\columncolor{Gray}}ccc>{\columncolor{Gray}}c>{\columncolor{Gray}}c}
\toprule
\midrule
\multicolumn{3}{c}{Training Data} & 
\multicolumn{4}{c}{R2R (Val Unseen)} & 
\multicolumn{4}{c}{REVERIE (Val Unseen)} & 
\multicolumn{4}{c}{\objnav-MP3D (Val)} \\ 

\cmidrule(r){1-3}
\cmidrule(r){4-7}
\cmidrule(r){8-11}
\cmidrule(r){12-15}

\multicolumn{1}{c}{R2R$^*$} & 
\multicolumn{1}{c}{REVERIE$^*$} & 
\multicolumn{1}{c}{MP3D$^*$} & 
\multicolumn{1}{c}{TL} & 
\multicolumn{1}{c}{NE$\downarrow$} & 
\multicolumn{1}{c}{SR$\uparrow$} & 
\multicolumn{1}{c}{SPL$\uparrow$} & 
\multicolumn{1}{c}{TL} & 
\multicolumn{1}{c}{NE$\downarrow$} & 
\multicolumn{1}{c}{SR$\uparrow$} & 
\multicolumn{1}{c}{SPL$\uparrow$} & 
\multicolumn{1}{c}{TL} & 
\multicolumn{1}{c}{NE$\downarrow$} & 
\multicolumn{1}{c}{SR$\uparrow$} & 
\multicolumn{1}{c}{SPL$\uparrow$} \\

\midrule\midrule
\cmark & & 
& 14.33 & 3.82 & 67 & 55
& \textcolor{textGray}{19.61} & \textcolor{textGray}{7.55} & \textcolor{textGray}{39} & \textcolor{textGray}{28}
& \textcolor{textGray}{15.30} & \textcolor{textGray}{4.69} & \textcolor{textGray}{55} & \textcolor{textGray}{24} \\
 & \cmark & 
& \textcolor{textGray}{17.55} & \textcolor{textGray}{6.22} & \textcolor{textGray}{42} & \textcolor{textGray}{32}
& 17.91 & 6.56 & 41 & 32
& \textcolor{textGray}{10.46} & \textcolor{textGray}{5.91} & \textcolor{textGray}{43} & \textcolor{textGray}{23} \\
 & & \cmark
& \textcolor{textGray}{20.76} & \textcolor{textGray}{8.55} & \textcolor{textGray}{16} & \textcolor{textGray}{9}
& \textcolor{textGray}{20.00} & \textcolor{textGray}{10.11} & \textcolor{textGray}{13} & \textcolor{textGray}{9}
& 22.17 & 3.67 & 68 & 29 \\
\cmark & & \cmark
& 14.03 & 4.01 & 64 & 55
& \textcolor{textGray}{15.22} & \textcolor{textGray}{7.78} & \textcolor{textGray}{38} & \textcolor{textGray}{31}
& 25.91 & 3.28 & 72 & 28 \\
 & \cmark & \cmark
& \textcolor{textGray}{19.17} & \textcolor{textGray}{7.13} & \textcolor{textGray}{34} & \textcolor{textGray}{26}
& 19.46 & 6.24 & 35 & 26
& 21.50 & 3.29 & 70 & 33 \\
\cmark & \cmark & \cmark
& 14.21 & 4.10 & 65 & 54
& 16.62 & 6.11 & 34 & 27
& 22.97 & 3.54 & 68 & 27 \\
\midrule
\end{tabular}
}
\caption{Comparison of single-run performance with a different mixture of training data for DuET~\cite{chen2022duet}. $*$ indicates utilizing Habitat-rendered images. \textcolor{Gray}{Numbers in gray} indicated zero-shot inference results on held-out datasets.}
\label{tab:multitasks}
\vspace{-10pt}
\end{table}

\paragraph{Data Transformation in Discrete Environment}
The original trajectories in \objnav are constructed as sequences of continuous viewpoint positions, averaging 243 steps per demonstration. Following Hong \etal~\cite{hong2022bridging}, we discretize the Habitat-MP3D~\cite{habitat19iccv} environment into a connectivity graph $ \mathcal{G}^* $. We match each viewpoint in the trajectory from Habitat-Web to the nearest nodes on $ \mathcal{G}^* $ based on Euclidean distance and merge sequentially repeated nodes. Disconnected paths and paths with an ending position more than $0.5 m$ away from the original endpoint are removed. This results in 58,803 trajectories with an average of 20 steps. Similarly, we transfer the data from the MP3D validation split to evaluate model performance in discrete environments.

\paragraph{Fine-grain Language Understanding Benefits Target-oriented Navigation}
We conduct multi-task training on DUET~\cite{chen2022duet} initialized from LXMERT~\cite{tan2019lxmert} using various data mixtures and report held-out inference results for datasets not included in the training. To address the visual gap between Habitat-rendered images and Matterport3D-rendered images, we map the R2R and REVERIE trajectories onto \( \mathcal{G}^* \) and employ Habitat-rendered images for this experiment. The results are presented in Table \ref{tab:multitasks}.

Our findings reveal several key insights: (1) Mixing training data across different tasks reduces performance compared to training on individual tasks requiring higher-level language understanding capacity. For example, in the R2R task, incorporating additional data results in a 2-3\% drop in success rate (SR), while in REVERIE, combining \objnav data leads to a substantial SR decrease of 6-7\%, even lower than the zero-shot performance achieved when training exclusively with R2R data (39\% SR vs. 35\% SR). (2) Training with fine-grained human-annotated instruction-trajectory pairs proves advantageous for \objnav, yielding a 2-4\% SR improvement. (3) Models trained exclusively on VLN data achieve strong zero-shot performance on \objnav (above 40\% SR); however, models trained only on \objnav data perform poorly on tasks that demand sophisticated language understanding (with only $\sim 15\%$ SR). Additionally, models trained solely on R2R (fine-grained VLN) data achieve 39\% SR on REVERIE (coarse-grained VLN) and 55\% SR on \objnav. This suggests that when the target is visible or only minor exploration is needed, R2R-trained models can successfully infer the location.
These observations lead to two main conclusions: 

\begin{enumerate}
    \item \textbf{Fine-grained language understanding improves target-oriented navigation}, as visual-semantic understanding is enhanced through learning vision-language alignment; however, models trained exclusively on target-oriented data lack the language comprehension required to follow complex instructions.
    \item \textbf{Training with simple data mixing is insufficient to achieve optimal performance} for tasks demanding both exploration and detailed instruction interpretation (coarse-grained VLN).
\end{enumerate}
\section{Mixture of Experts for Versatile Language-guided Visual Navigation}

\begin{figure*}[h]
  \centering
   \includegraphics[width=0.8\linewidth]{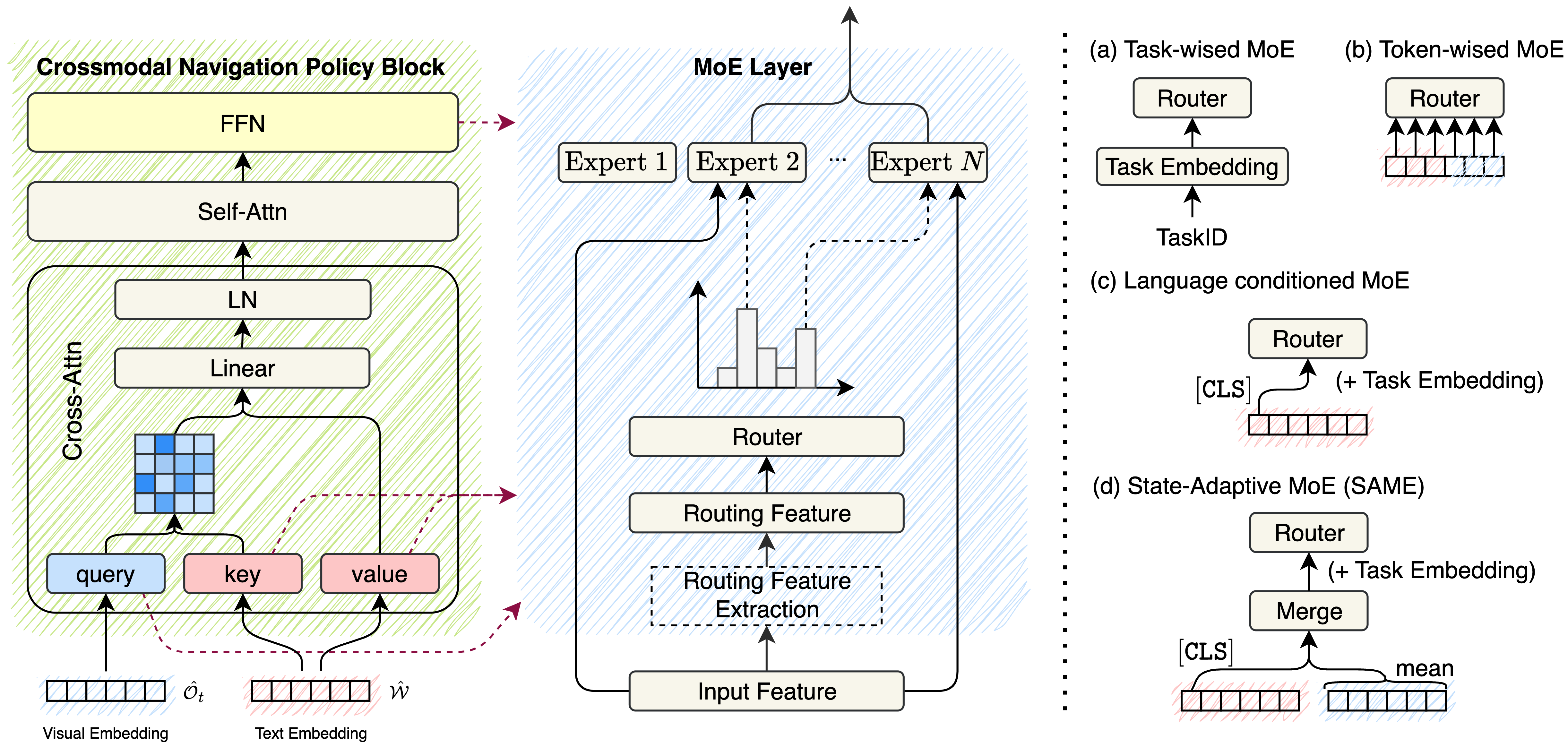}

   \caption{
   Illustration of MoE position and experts' routing methods. \ours~routing based on multimodal features from visual observations and language instructions allows the agent to dynamically adapt to environmental visual changes.
   }
   \label{fig:pipeline}
\end{figure*}

The insights from Section \ref{sec:conflicts} motivate the need for a method to manage conflicts that arise during multi-task learning. To address this, we propose a new State-Adaptive Mixture of Experts (SAME) approach. SAME employs multiple specialized expert networks $\mathcal{E}=\{f_1, \cdots, f_N\}$ which could be switched during each step in a navigation episode conditioned on the agent's state by a routing mechanism $\mathcal{R}$. In this way, we differentiate the learning of distinct navigation skills, such as exploration and instruction-following, while facilitating the sharing of common navigational knowledge like visual semantic understanding.

\subsection{MoE Formulation}

A subset of experts is activated in a sparsely gated MoE layer during each forward pass. The router predicts the probability of each expert being assigned:
\begin{align}
    \mathcal{P}(\mathbf{x}_r) &= \text{Softmax}\left(\mathcal{R}\left(\mathbf{x}_r\right)\right)\\
    \mathcal{R}(\mathbf{x}_r) &= W \mathbf{x}_r
\end{align}
where $\mathbf{x}_r$ is the routing feature extracted from the input $\mathbf{x}$, $W \in \mathbb{R}^{d \times N}$ is a trainable layer, $d$ is the hidden dimension and $N$ is the number of experts. The weighted sum of the outputs from experts with top-$k$ routing scores noted as set $\mathcal{T}$ is computed as the output:
\begin{equation}
    \text{MoE}(\mathbf{x}, \mathbf{x}_r) = \sum_{i\in \mathcal{T}}  \mathcal{P}\left(\mathbf{x}_r\right)_i f_i(\mathbf{x})
\end{equation}
Typically a load balancing loss~\cite{fedus2022switch} is implemented to encourage an even distribution of input across the $N$ experts:
\begin{align}
    \mathcal{L}_\text{balance} &= N \sum_{i=0}^N \mathcal{F}_i\mathcal{D}_i \label{equ:balance_loss} \\
    \mathcal{F} &= \frac{1}{K} \sum_{i=1}^{N} \mathbf{1} \left\{ \arg\max \mathcal{P}(\mathbf{x}_r) = i \right\}, \\
    \mathcal{D} &= \frac{1}{K} \sum_{i=1}^{K} \mathcal{P}(\mathbf{x}_r)_i.
\end{align}
where $\mathcal{F}$ represents the fraction of inputs processed by each expert $f_i$, and $\mathcal{D}$ represents the fraction of router probability allocated for expert $f_i$.

\noindent\textbf{Task-wised MoE and Token-wised MoE}
Recently, most research on Mixture of Experts (MoE) has focused on transformer-based Large Language Models (LLMs), where MoE operates at the token level to process each individual input token, as illustrated in Figure \ref{fig:pipeline}. Concurrently, MoE is also extensively explored in computer vision multi-task learning, where expert modules are routed at the task level. These two routing features are formulated as follows:
\begin{align}
    \mathbf{x}^\text{token}_r &= \mathbf{x}_i \\
    \mathbf{x}^\text{task}_r &= E^\text{task} = W_t T
\end{align}
where $\mathbf{x}_i$ is the $i$-th token in input $\mathbf{x}$, and $E^\text{task}$ is the task embedding for task with index $T$.

Beyond assigning each task $T$ to specific experts through a hard assignment, it might be helpful for an agent to select appropriate navigation skills based on the language instruction. To this end, we formulate a language-aware expert selection mechanism, represented as follows:
\begin{equation}
    \mathbf{x}^\text{text}_r = \hat{\mathcal{W}}^\texttt{CLS}
\end{equation}
where $\hat{\mathcal{W}}^\texttt{CLS}$ denotes the $[\texttt{CLS}]$ token for text feature $\hat{\mathcal{W}}$.

\subsection{State-Adaptive Experts Selection}
Our initial experiments with the above token-wise and task-wise MoE in multi-task navigation learning yielded suboptimal results (to be presented in \S\ref{sec:moe_type}), prompting us to reconsider a more feasible MoE formulation for the sequential decision-making process in navigation. We observed that the density of language information the agent receives -- which must align with visual observations to determine an action -- can vary significantly across different timesteps in different tasks.
In other words, the agent's state interpretation should be inherently generalizable to address distinct navigation problems.
In light of this, we introduce \ours, a multimodal State-Adaptive expert selection mechanism:
\begin{equation}
    \mathbf{x}^\text{multi}_r = W_m \left[\frac{1}{L}\sum_{i=0}^L\hat{\mathcal{O}_t}; \hat{\mathcal{W}}^\texttt{CLS}\right]
\end{equation}
where a linear layer $W_m \in \mathbb{R}^{2h \times h}$ is implemented to merge the concatenation of the mean visual feature $\frac{1}{L}\sum_{i=0}^L\hat{\mathcal{O}_t}$ over different views and the text $[\texttt{CLS}]$ token $\hat{\mathcal{W}}^\texttt{CLS}$.

Besides, we also investigate the effect of adding task information for expert selection:
\begin{align}
    \mathbf{x}^\text{text\_task}_r &= \mathbf{x}^\text{text}_r + E^\text{task} \\
    \mathbf{x}^\text{multi\_task}_r &= \mathbf{x}^\text{multi}_r + E^\text{task}
\end{align}

\subsection{Comparison on MoE Routing}
\label{sec:moe_type}
Following the above, we can see the key to distinguishing specific and shared navigational knowledge learning is learning the routing mechanism in MoE. In this section, we investigate the routing strategies for a versatile language-guided navigation policy.

\paragraph{Experiment Setup}
\begin{table}[t]
\centering

\resizebox{\columnwidth}{!}{
\definecolor{Gray}{gray}{0.94}

\begin{tabular}{lcc>{\columncolor{Gray}}c>{\columncolor{Gray}}ccc>{\columncolor{Gray}}c>{\columncolor{Gray}}ccc>{\columncolor{Gray}}c>{\columncolor{Gray}}c}
\toprule
\midrule
\multicolumn{1}{c}{\multirow{2}{*}{Routing Condition}} & 
\multicolumn{4}{c}{R2R (Val Unseen)} & 
\multicolumn{4}{c}{REVERIE (Val Unseen)} & 
\multicolumn{4}{c}{\objnav-MP3D (Val)} \\ 

\cmidrule(r){2-5}
\cmidrule(r){6-9}
\cmidrule(r){10-13}

\multicolumn{1}{c}{} & 
\multicolumn{1}{c}{TL} & 
\multicolumn{1}{c}{NE$\downarrow$} & 
\multicolumn{1}{c}{SR$\uparrow$} & 
\multicolumn{1}{c}{SPL$\uparrow$} & 
\multicolumn{1}{c}{TL} & 
\multicolumn{1}{c}{NE$\downarrow$} & 
\multicolumn{1}{c}{SR$\uparrow$} & 
\multicolumn{1}{c}{SPL$\uparrow$} & 
\multicolumn{1}{c}{TL} & 
\multicolumn{1}{c}{NE$\downarrow$} & 
\multicolumn{1}{c}{SR$\uparrow$} & 
\multicolumn{1}{c}{SPL$\uparrow$} \\

\midrule\midrule
w/o MoE
& 11.76 & 2.98 & 73.09 & \textbf{65.79}
& 13.17 & 5.90 & 40.39 & 35.40
& 16.13 & 3.16 & 72.31 & 42.72 \\
\midrule
\rowcolor{Cerulean!20}\multicolumn{13}{l}{\emph{Token-wised MoE}~\cite{shazeer2017outrageously}:}\\
Token Embedding
& 13.28 & 2.98 & 73.99 & 64.81
& 16.78 & 5.60 & 43.45 & 35.40
& 15.69 & 3.04 & \textbf{74.38} & \textbf{44.97} \\
\rowcolor{Cerulean!20}\multicolumn{13}{l}{\emph{Task-wised MoE}:}\\
Task Embedding
& 12.98 & 3.13 & 72.84 & 64.81
& 14.90 & 5.71 & 43.71 & 36.58
& 14.96 & 3.17 & 71.13 & 43.88 \\
Text \texttt{[CLS]}
& 14.45 & 2.92 & \textbf{74.67 }& 64.12
& 19.22 & 5.46 & 45.50 & 36.56
& 15.63 & 3.11 & 71.32 & 42.75 \\
\quad w/ Task Embedding
& 15.80 & 2.95 & 73.61 & 62.80
& 21.50 & 5.44 & 43.85 & 34.10
& 18.26 & 3.00 & 72.40 & 39.98 \\
\rowcolor{Cerulean!20}\multicolumn{13}{l}{\emph{State-Adaptive MoE (ours)}:}\\
\ours
& 13.51 & \textbf{2.90} & 73.69 & \textbf{64.92}
& 16.32 & \textbf{5.38} & \textbf{45.67} & \textbf{37.95}
& 15.60 & 3.10 & 71.43 & 43.39 \\
\quad w/ Task Embedding
& 14.00 & 2.89 & 74.50 & 64.66
& 17.65 & 5.66 & 42.32 & 33.61
& 17.20 & \textbf{2.86} & 73.39 & 42.07 \\
\quad w/o Pretrain
& 14.71 & 4.66 & 59.05 & 49.02
& 14.19 & 6.26 & 38.43 & 32.40
& 19.93 & 3.08 & 70.94 & 38.48 \\
\bottomrule
\end{tabular}
}
\caption{Comparison of single-run performance with different MoE routing conditions on R2R, REVERIE, and \objnav-MP3D.}
\vspace{-5pt}
\label{tab:moe_type}
\end{table}
We conduct contrastive experiments to compare the routing mechanism, Matterport3D-rendered images~\cite{anderson2018r2r} are used for R2R and REVERIE. MoE experts are deployed to the visual queries (will be discussed in Section \ref{sec:moe_position}), and results are shown in Table \ref{tab:moe_type}.
As discussed in Section \ref{sec:conflicts}, we demonstrate that learning fine-grained language understanding can enhance target-oriented navigation learning, motivating us to take advantage of the powerful vision-language-action pre-train conducted in VLN~\cite{hao2020prevalent, guhur2021airbert, majumdar2020improving, chen2021hamt, chen2022duet, qiao2023hop+, wang2023scaling}. Throughout the subsequent experiments, we initialized our model from ScaleVLN~\cite{wang2023scaling}, which scales the VLN data from 14K to 4.9M by generating synthetic data to perform pertaining.

\paragraph{VLN Pretrain Benefits Multiple Navigation Task Learning}
As shown in Table \ref{tab:moe_type}, there is a significant improvement on all tasks when initializing with VLN pre-trained weights compared to directly performing multi-task tuning on \ours~ initialized from general vision-language pretrain LXMERT~\cite{tan2019lxmert} (w/o Pretrain), featured by a $\sim 15\%$ SR increase on R2R. 

\noindent\textbf{\ours~Facilitates Navigation Skill Sharing}
Comparing different MoE routing types, we highlight a significant improvement of $\sim 3\%$ SR and $\sim 2\%$ SPL in \ours~routing on REVERIE. We hypothesize that this improvement is due to the nature of the coarse-grained VLN task, where the agent must alternate between exploration, in cases where language guidance lacks detail, and adhering closely to language instructions when precisely localizing the target. \ours~routing enables a flexible selection of experts to manage these distinct navigation behaviors based on the current observation and language input, allowing the agent to learn transferable knowledge across tasks. This flexibility also results in the \ours~agent achieving the highest average SPL of 48.31\% across all tasks.

Compared to the multi-task-tuned DUET (w/o MoE) under the same experimental conditions, all MoE methods show a significant performance increase (3-5\% SR) on the REVERIE task, with no notable performance drops on other tasks. This aligns with our findings in Section \ref{sec:conflicts}, suggesting that coarse-grained VLN performance is influenced when co-trained with other language instruction types, likely due to model overfitting to the R2R task. Our proposed MoE approach addresses this issue effectively.

Furthermore, we examine hard assignments with specific experts dedicated to different tasks by directly routing through task embeddings. This approach results in performance drops across all tasks compared to other routing methods. 
Additionally, incorporating task embeddings into $\mathbf{x}^\text{text}_r$ and $\mathbf{x}^\text{multi}_r$ reduces performance, indicating that routing based on samples from different tasks provides strong but ineffective bias, preventing the router from effectively learning which experts to select based on the instruction or observation, thus impeding the development of shared knowledge across tasks. 

\subsection{Which Part of the Navigation Policy Learns Different Navigation Behaviour?}
\label{sec:moe_position}
\begin{table}[t]
\centering

\resizebox{\columnwidth}{!}{
\definecolor{Gray}{gray}{0.94}

\begin{tabular}{lcc>{\columncolor{Gray}}c>{\columncolor{Gray}}ccc>{\columncolor{Gray}}c>{\columncolor{Gray}}ccc>{\columncolor{Gray}}c>{\columncolor{Gray}}c}
\toprule
\midrule
\multicolumn{1}{c}{\multirow{2}{*}{MoE Experts Position}} & 
\multicolumn{4}{c}{R2R (Val Unseen)} & 
\multicolumn{4}{c}{REVERIE (Val Unseen)} & 
\multicolumn{4}{c}{\objnav-MP3D (Val)} \\ 

\cmidrule(r){2-5}
\cmidrule(r){6-9}
\cmidrule(r){10-13}

\multicolumn{1}{c}{} & 
\multicolumn{1}{c}{TL} & 
\multicolumn{1}{c}{NE$\downarrow$} & 
\multicolumn{1}{c}{SR$\uparrow$} & 
\multicolumn{1}{c}{SPL$\uparrow$} & 
\multicolumn{1}{c}{TL} & 
\multicolumn{1}{c}{NE$\downarrow$} & 
\multicolumn{1}{c}{SR$\uparrow$} & 
\multicolumn{1}{c}{SPL$\uparrow$} & 
\multicolumn{1}{c}{TL} & 
\multicolumn{1}{c}{NE$\downarrow$} & 
\multicolumn{1}{c}{SR$\uparrow$} & 
\multicolumn{1}{c}{SPL$\uparrow$} \\

\midrule\midrule
Feed Forward
& 13.18 & 3.08 & 73.44 & 64.86
& 16.02 & 5.62 & 42.77 & 35.28
& 16.02 & 3.09 & 71.37 & 42.24 \\
Visual Query
& 13.51 & 2.90 & 73.69 & \textbf{64.92}
& 16.32 & 5.38 & \textbf{45.67} & \textbf{37.95}
& 15.60 & 3.10 & 71.43 & \textbf{43.39} \\
Textual Key \& Value
& 15.58 & 2.82 & \textbf{75.35} & 63.85
& 20.30 & 5.36 & 45.61 & 34.75
& 17.57 & 3.10 & \textbf{72.17} & 42.67 \\

\midrule
\end{tabular}
}
\caption{Comparison of single-run performance with different MoE experts' positions on R2R, REVERIE, and \objnav-MP3D.}
\label{tab:moe_position}
\vspace{-5pt}
\end{table}
Since the initial application of MoE in transformer architectures~\cite{fedus2022switch, lepikhin2020gshard, zoph2022st}, MoE has acted as an enhancement for Feed-Forward Network (FFN) modules within these models. Concurrently, some studies have integrated multi-head attention layers with MoE to further enhance performance while managing computational costs. In this section, we analyze the impact of MoE applied at different components of the transformer model, specifically focusing on the FFN, visual queries $W_q$, textual key $W_k$, and value $W_v$ with \ours~under the same experiment setup described in Section \ref{sec:moe_type}. The results are shown in Table \ref{tab:moe_position}.

MoE applied to different components yields varying performance across tasks. Notably, the best overall performance is observed when applying MoE to the visual query, achieving the highest SPL on all tasks with fewer parameters. This suggests that utilizing MoE at the visual query level within the cross-attention layer is particularly effective. This effectiveness likely stems from the multimodal policy’s control over diverse navigation behaviors within the cross-attention layer, where specialized visual query experts allow the agent to more accurately determine the next action by adjusting attention scores over visual embeddings from multiple viewpoints. While FFN-based MoE enhances performance compared to non-MoE models, it is surpassed by MoE configurations that integrate with attention layers, highlighting the crucial role of cross-modal attention in successful action selection.

In summary, the experimental results indicate that employing MoE with visual query experts and routing based on multimodal features from visual observations and language instructions (\ours) allows the agent to dynamically adapt to environmental visual changes while staying aligned with language guidance, thereby enhancing robust performance across different language-guided navigation tasks.
\section{Experiments}
\begin{table*}[t]
\centering
\resizebox{\textwidth}{!}{
\definecolor{Gray}{gray}{0.94}
\begin{tabular}{lcccccccccccccccccc}
\toprule
\midrule
\multicolumn{1}{c}{\multirow{3}{*}{Methods}} 
& \multicolumn{2}{c}{CVDN} & \multicolumn{2}{c}{RxR-EN} & \multicolumn{4}{c}{R2R} & \multicolumn{4}{c}{SOON} & \multicolumn{4}{c}{REVERIE} & \multicolumn{2}{c}{\objnav-MP3D} \\ 
\cmidrule(r){2-3}
\cmidrule(r){4-5}
\cmidrule(r){6-9}
\cmidrule(r){10-13}
\cmidrule(r){14-17}
\cmidrule(r){18-19}
 & \multicolumn{1}{c}{Val} & \multicolumn{1}{c}{Test} 
 & \multicolumn{2}{c}{Val unseen} 
 & \multicolumn{2}{c}{Val unseen} & \multicolumn{2}{c}{Test unseen} 
 & \multicolumn{2}{c}{Val unseen} & \multicolumn{2}{c}{Test unseen} 
 & \multicolumn{2}{c}{Val unseen} & \multicolumn{2}{c}{Test unseen}
 & \multicolumn{2}{c}{Val} \\
 \cmidrule(r){2-2} \cmidrule(r){3-3} 
 \cmidrule(r){4-5} 
 \cmidrule(r){6-7} \cmidrule(r){8-9} 
 \cmidrule(r){10-11} \cmidrule(r){12-13}
 \cmidrule(r){14-15} \cmidrule(r){16-17}
 \cmidrule(r){18-19}
 & GP $\uparrow$ & GP $\uparrow$ 
 & SR $\uparrow$ & nDTW $\uparrow$ 
 & SR $\uparrow$ & SPL $\uparrow$ & SR $\uparrow$ & SPL $\uparrow$ 
 & SR $\uparrow$ & SPL $\uparrow$ & SR $\uparrow$ & SPL $\uparrow$ 
 & SR $\uparrow$ & SPL $\uparrow$ & SR $\uparrow$ & SPL $\uparrow$ 
 & SR $\uparrow$ & SPL $\uparrow$ \\ 
\midrule
\midrule
\rowcolor{Cerulean!20}\multicolumn{19}{l}{\emph{Separate Model for Each Task}:}\\
SF~\cite{fried2018speaker}
& - & - 
& - & -
& 36 & - & 35 & 28 
& - & - & - & -
& - & - & - & - 
& - & - \\
RCM~\cite{wang2019reinforced}
& - & - 
& - & -
& 43 & - & 43 & 38 
& - & - & - & - 
& 9.3 & 7.0 & 7.8 & 6.7
& - & - \\
EnvDrop~\cite{tan2019envdrop}
& - & - 
& - & -
& 52 & 48 & 51 & 47 
& - & - & - & -
& - & - & - & - 
& - & - \\
PREVALENT~\cite{hao2020prevalent}
& 3.15 & 2.44 
& - & -
& 58 & 53 & 54 & 51

& - & - & - & -
& - & - & - & - 
& - & - \\
\vlnbert~\cite{hong2020recurrent}
& - & - 
& - & -
& 63 & 57 & 63 & 57

& - & - & - & - 
& 25.5 & 21.1 & 24.6 & 19.5 

& - & - \\
HAMT~\cite{chen2021history}
& 5.13 & 5.58 
& 56.4 & 63.0 
& 66 & 61 & 65 & 60

& - & - & - & - 
& 33.0 & 30.2 & 30.4 & 26.7 

& - & - \\
HOP+~\cite{qiao2023hop+}
& - & - 
& - & -
& 67 & 61 & 66 & 60

& - & - & - & - 
& 36.1 & 31.1 & 33.8 & 28.2

& - & - \\
DUET~\cite{chen2022duet}
& - & - 
& - & -
& 72 & 60 & 69 & 59

& 36.3 & 22.6 & 33.4 & 21.4 
& 47.0 & 33.7 & 52.5 & 36.1 

& - & - \\
AutoVLN~\cite{chen2022hm3dlearning}
& - & - 
& - & -
& - & - & - & -

& 41.0 & 30.7 & 40.4 & 27.9 
& 55.9 & 40.9 & 55.2 & 38.9 

& - & - \\
BEVBert~\cite{an2022bevbert}
& - & - 
& 66.7 & 69.6
& 75 & 64 & 73 & 62 
& - & - & - & - 
& 51.8 & 36.4 & 52.8 & 36.4 
& - & - \\
GridMM~\cite{wang2023gridmm}
& - & - 
& - & -
& 75 & 64 & 73 & 62 
& - & - & - & -
& - & - & - & - 
& - & - \\
VER~\cite{liu2024volumetric}
& - &  
& - & -
& 76 & 65 & 76 & 66

& - & - & - & - 
& 56.0 & 39.7 & 56.8 & 38.8

& - & - \\
GOAT~\cite{wang2024vision}
& - & - 
& 68.2 & 66.8
& 78 & 68 & 75 & 65

& 40.4 & 28.1 & 40.5 & 25.2 
& 53.4 & 36.7 & 57.7 & 40.5 

& - & - \\
ScaleVLN~\cite{wang2023scaling}
& 6.12 & 6.97 
& - & -
& 79 & 70 & 77 & 68

& - & - & - & - 
& 57.0 & 41.8 & 56.1 & 39.5 

& - & - \\
\midrule
\rowcolor{Cerulean!20}\multicolumn{19}{l}{\emph{Unified Model for All Tasks}:}\\
MT-RCM~\cite{wang2020environment}
& 4.65 & 3.91 
& - & -
& 47 & 41 & 45 & 40

& - & - & - & - 
& - & - & - & - 
& - & - \\
NaviLLM~\cite{zheng2023towards}
& 6.16 & \textbf{7.90} 
& - & - 
& 67 & 59 & 68 & 60

& \textbf{38.3} & \textbf{29.2} & 35.0 & 26.3
& 42.2 & 35.7 & 39.8 & 32.3

& - & - \\
ScaleVLN$\dagger$
& 5.93 & -
& 46.7 & 49.7 
& 76 & \textbf{67} & - & - 
& 33.2 & 25.4 & - & - 
& 41.9 & 34.4 & - & - 
& 72.3 & \textbf{43.4} \\
\ours~(ours)
& \textbf{6.94} & 7.07
& \textbf{50.5} & \textbf{51.2}
& \textbf{76} & 66 & \textbf{74} & \textbf{64} 
& 36.1 & 25.4 & \textbf{38.2} & \textbf{27.1} 
& \textbf{46.4} & \textbf{36.1} & \textbf{48.6} & \textbf{37.1} 
& \textbf{76.3} & 42.7 \\
\bottomrule
\end{tabular}}
\caption{Agents performance across all tasks in the discrete environment~\cite{anderson2018r2r}. $\dagger$ indicates our implementation of multi-task tuning. Note that existing methods tailored for \objnav-MP3D are proposed in continuous environments, which will be evaluated in Table~\ref{tab:continuous} below.}
\label{tab:discrete}
\end{table*}
In this section, we conduct multi-task tuning based on the previous discussion. We conduct a thorough evaluation of \ours~across seven major navigation benchmarks, complemented by a series of ablation studies on training details to establish best practices for effective multi-task training in language-guided navigation.

\vspace{5pt}
\noindent\textbf{Datasets}
Besides the R2R, REVERIE, and \objnav-MP3D, we include 4 other datasets for evaluation.
\begin{itemize}
    \item RxR-EN~\cite{anderson2020rxr}: English split of the RxR dataset, which contains longer instructions compared to R2R and non-shortest trajectory from starting point to ending point.
    \item CVDN~\cite{thomason2020cvdn} requires the agent to comprehend the conversation history and infer the correct next actions based on the dialogue context. For evaluation, we use the standard metric, Goal Progress (GP), which calculates the average difference between the completed trajectory length and the remaining distance to the goal.
    \item SOON~\cite{zhu2021soon}: Similar to REVERIE, the instructions describe target rooms and objects, with an average length of 47 words. The expert paths vary in length from 2 to 21 steps, with an average of 9.5 steps.
    \item R2R-CE~\cite{krantz2020beyond}: Transfering the discrete trajectories in R2R to continuous 3D scans rendered by Habitat~\cite{savva2019habitat}, allowing an agent to navigate freely in open space while requiring interaction with obstacles.
\end{itemize}

\vspace{5pt}
\noindent\textbf{Implementation Details}
We build upon the DUET architecture, and replace all the visual query layers in the cross-attention with MoE layers. We adopt \ours~routing for each MoE layer.
We initialize from the pre-trained weights of ScaleVLN~\cite{wang2023scaling} and utilize CLIP ViT-B/16~\cite{radford2021clip} as the visual encoder.
We use ScaleVLN, R2R, RxR-EN, CVDN, REVERIE, SOON, and Habitat-Web as the training data with a sampling ratio of 10:1:1:1:1:1:2 without mixing different data in a batch.
We follow~\cite{chen2022duet, an2022bevbert, kamath2023new} to utilize DAgger~\cite{ross2011dagger} algorithm to obtain interactive supervision from the simulator. The training objective $\mathcal{L} = \mathcal{L}_\text{DAG} + \lambda \mathcal{L}_\text{balance}$ is the combination of DAgger loss and MoE load balancing loss in Equation \ref{equ:balance_loss}, balanced by coefficient $\lambda = 0.8$.
The model is fine-tuned using AdamW optimizer~\cite{loshchilov2018decoupled} with a learning of $1 \times 10^{-5}$ for 20k iterations with a batch size of 16 on a single 80G NVIDIA A100 GPU.

\subsection{Comparison with State-of-the-Art Models}
\noindent\textbf{Comparison in Discrete Environment}
We first compare our method with current SoTA methods in the discrete MP3D environment~\cite{anderson2018r2r}. The datasets are organized in Table \ref{tab:discrete} by language instruction granularity and complexity, ranging from fine-grained and complex to coarse-grained and simple, from left to right.
We classify previous methods into two categories, the first one is a separate model for each task, which fine-tunes a distinct set of parameters for each downstream task. The second one is a unified model for all tasks, which includes our baseline methods:
\begin{itemize}
    \item MT-RCM~\cite{wang2020environment}: This method performs multi-task training on R2R and CVDN using the RCM model~\cite{wang2019reinforced}.
    \item NaviLLM~\cite{zheng2023towards}: A hybrid model of DUET and Large Language Model (LLM) that employs a stronger vision encoder, EVA-CLIP-02-Large~\cite{fang2023eva}, and replaces the 12-layer cross-modal encoder in DUET with a LLM, Vicuna-7B~\cite{vicuna2023}, to facilitate multi-task instruction tuning.
    \item ScaleVLN$\dagger$: We conduct multi-task tuning on DUET~\cite{chen2022duet} with ScaleVLN~\cite{wang2023scaling} initilization. This serves as a baseline for multi-task tuning on DUET without using MoE.
\end{itemize}

As shown in Table \ref{tab:discrete}, \ours~achieve State-of-the-Art performance on multiple benchmarks with a unified model. Compared to separate models for each task, \ours~achieve SoTA performance on CVDN with a 7.07 GP on test 13\% GP increase on validation compared to ScaleVLN. \ours~performs at the same level as VER and GOAT comparing the SPL on R2R and REVERIE. Compared to other unified models, \ours~outperform the baseline multi-task tuned ScaleVLN on all tasks, with an average of 3\% SR increase among all tasks. \ours~peform significantly better than NaviLLM on R2R and REVERIE.

\noindent\textbf{Comparison in Continuous Environment}
\begin{table}[t]
\centering
\small

\resizebox{0.8\columnwidth}{!}{
\definecolor{Gray}{gray}{0.94}
\definecolor{textGray}{gray}{0.5}
\begin{tabular}{lccccccccccc}
\toprule
\midrule
\multicolumn{1}{c}{\multirow{3}{*}{Methods}} 
& \multicolumn{3}{c}{R2R-CE (Val unseen)} & \multicolumn{2}{c}{MP3D (Val)}\\ 
\cmidrule(r){2-4}
\cmidrule(r){5-6}
 & NE $\downarrow$ & SR $\uparrow$ & SPL$\uparrow$
 & SR $\uparrow$ & SPL $\uparrow$ \\ 
\midrule
\midrule
NaVid~\cite{zhang2024navid}
& 5.47 & 37 & 36 
& - & - \\
ScaleVLN~\cite{wang2023scaling}
& 4.80 & 55 & 51 
& - & - \\
ETPNav~\cite{an2023etpnav}
& 4.71 & 57 & 49 
& - & - \\
BEVBert~\cite{an2022bevbert}
& \textbf{4.57} & \textbf{59} & \textbf{50} 
& - & - \\
\midrule
SemExp~\cite{chaplot2020object}
& - & - & - 
& 28 & 11 \\
PONI~\cite{ramakrishnan2022poni}
& - & - & - 
& 32 & 12\\
Habitat-Web~\cite{ramrakhya2022habitatweb}
& - & - & - 
& 35 & 10\\
\midrule
\ours~(ours)
& \textcolor{textGray}{5.31} & \textcolor{textGray}{47} & \textcolor{textGray}{38} 
& \textbf{\textcolor{textGray}{43}} & \textbf{\textcolor{textGray}{21}} \\
\midrule\bottomrule
\end{tabular}
}
\caption{Comparison with previous methods in the continuous environment~\cite{savva2019habitat}. We report the \textcolor{Gray}{zero-shot inference results} of SAME using the same checkpoint from Table~\ref{tab:discrete}.}
\label{tab:continuous}
\end{table} 
We further compare our method with current SoTA methods in the continuous Habitat environment~\cite{savva2019habitat}. We take the same model reported in Table \ref{tab:discrete} and examine the zero-shot inference results on held-out datasets in continuous environment. We follow Hong \etal~\cite{hong2022bridging} and deploy the waypoint predictor to bridge the gap between discrete and continuous. As shown in Table \ref{tab:continuous}, \ours~surpass Habitat-Web, which performs imitation learning on \objnav human demonstration, when zero-shot transferring to the continuous environment. This aligned with our findings in \S\ref{sec:conflicts} that fine-grain language understanding improves target-oriented navigation.

\subsection{Ablation Studies}
We conduct a series of experiments to establish best practices for training with multiple navigation tasks. These ablations are performed using the same experimental setup as described in Section \ref{sec:moe_type}.

\noindent\textbf{{Effect of Training Schema}}
\begin{table}[t]
\centering

\resizebox{\columnwidth}{!}{
\definecolor{Gray}{gray}{0.94}

\begin{tabular}{cccc>{\columncolor{Gray}}c>{\columncolor{Gray}}ccc>{\columncolor{Gray}}c>{\columncolor{Gray}}ccc>{\columncolor{Gray}}c>{\columncolor{Gray}}c}
\toprule
\midrule
\multicolumn{1}{c}{\multirow{2}{*}{Algorithm}} & 
\multicolumn{1}{c}{\multirow{2}{*}{Batch}} & 
\multicolumn{4}{c}{R2R (Val Unseen)} & 
\multicolumn{4}{c}{REVERIE (Val Unseen)} & 
\multicolumn{4}{c}{\objnav-MP3D (Val)} \\ 

\cmidrule(r){3-6}
\cmidrule(r){7-10}
\cmidrule(r){11-14}

\multicolumn{2}{c}{} & 
\multicolumn{1}{c}{TL} & 
\multicolumn{1}{c}{NE$\downarrow$} & 
\multicolumn{1}{c}{SR$\uparrow$} & 
\multicolumn{1}{c}{SPL$\uparrow$} & 
\multicolumn{1}{c}{TL} & 
\multicolumn{1}{c}{NE$\downarrow$} & 
\multicolumn{1}{c}{SR$\uparrow$} & 
\multicolumn{1}{c}{SPL$\uparrow$} & 
\multicolumn{1}{c}{TL} & 
\multicolumn{1}{c}{NE$\downarrow$} & 
\multicolumn{1}{c}{SR$\uparrow$} & 
\multicolumn{1}{c}{SPL$\uparrow$} \\

\midrule\midrule
Imitation
& Mix
& 13.77 & 3.82 & 65.51 & 56.57
& 19.02 & 5.79 & 35.64 & 28.15
& 31.31 & 2.64 & \textbf{75.83} & 26.77 \\
Imitation
& Sequential
& 9.35 & 4.23 & 61.09 & 58.88
& 9.24 & 7.37 & 28.37 & 26.42
& 26.57 & 3.51 & 71.70 & 24.06 \\
\midrule
DAgger
& Mix
& 16.09 & 4.07 & 62.36 & 52.18
& 25.05 & 5.89 & 30.45 & 21.91
& 27.57 & 2.81 & 74.85 & 31.70 \\
DAgger
& Sequential
& 13.51 & 2.90 & \textbf{73.69} & \textbf{64.92}
& 16.32 & 5.38 & \textbf{45.67} & \textbf{37.95}
& 15.60 & 3.10 & 71.43 & \textbf{43.39} \\
\midrule
\end{tabular}
}
\caption{Comparison of single-run performance with different training schema on R2R, REVERIE, and \objnav-MP3D.}
\label{tab:schema}
\end{table}
We investigate different training schemas for SAME, specifically the training algorithm and data mixing strategy. For the training algorithm, we compare training with imitation learning only on teacher actions~\cite{anderson2018r2r} to training with DAgger, where at each time step, an agent performs an action sampled from the predicted probability of its action space and minimizes the loss between the sampled action and the ground truth. This method allows an agent to learn from paths that cover wide space and reduces the exposure bias caused by teacher forcing [46]. For the data mixing strategy, we investigate mixing different datasets in the same batch versus sampling different data for training.
As shown in Table \ref{tab:schema}, training with DAgger significantly improves the performance when sequentially sampling data. Compare row 4 to row 2 in Table \ref{tab:schema}, 12.6\% SR and 6.04\% SPL increase on R2R, 17.3\% SR and 11.53\% SPL increase on REVERIE, 19.33\% SPL increase on \objnav-MP3D is observed. We also noticed that mixing up training data in the same batch would significantly affect DAgger training, comparing row 4 to row 3. This indicates utilizing Dagger without mixing up training data in the same batch is the best practice. 

\noindent\textbf{Effect of MoE Routing Balance Coefficient}
\begin{table}[t]
\centering

\resizebox{\columnwidth}{!}{
\definecolor{Gray}{gray}{0.94}

\begin{tabular}{lcc>{\columncolor{Gray}}c>{\columncolor{Gray}}ccc>{\columncolor{Gray}}c>{\columncolor{Gray}}ccc>{\columncolor{Gray}}c>{\columncolor{Gray}}c}
\toprule
\midrule
\multicolumn{1}{c}{\multirow{2}{*}{$\lambda$}} & 
\multicolumn{4}{c}{R2R (Val Unseen)} & 
\multicolumn{4}{c}{REVERIE (Val Unseen)} & 
\multicolumn{4}{c}{\objnav-MP3D (Val)} \\ 

\cmidrule(r){2-5}
\cmidrule(r){6-9}
\cmidrule(r){10-13}

\multicolumn{1}{c}{} & 
\multicolumn{1}{c}{TL} & 
\multicolumn{1}{c}{NE$\downarrow$} & 
\multicolumn{1}{c}{SR$\uparrow$} & 
\multicolumn{1}{c}{SPL$\uparrow$} & 
\multicolumn{1}{c}{TL} & 
\multicolumn{1}{c}{NE$\downarrow$} & 
\multicolumn{1}{c}{SR$\uparrow$} & 
\multicolumn{1}{c}{SPL$\uparrow$} & 
\multicolumn{1}{c}{TL} & 
\multicolumn{1}{c}{NE$\downarrow$} & 
\multicolumn{1}{c}{SR$\uparrow$} & 
\multicolumn{1}{c}{SPL$\uparrow$} \\

\midrule\midrule
0.2
& 12.63 & 2.91 & 73.61 & \textbf{65.73}
& 15.47 & 5.55 & 42.86 & 35.22
& 15.25 & 3.13 & 71.79 & 43.51 \\
0.5
& 13.77 & 2.89 & \textbf{74.29} & 65.00
& 19.45 & 5.56 & 44.70 & 34.89
& 18.10 & 3.05 & 72.26 & 40.50 \\
0.8
& 13.51 & 2.90 & 73.69 & 64.92
& 16.32 & 5.38 & 45.67 & \textbf{37.95}
& 15.60 & 3.10 & 71.43 & 43.39 \\
1.0
& 13.56 & 3.02 & 73.56 & 64.52
& 16.19 & 5.24 & \textbf{45.81} & 37.89
& 18.11 & 2.67 & \textbf{74.66} & \textbf{44.81}\\

\midrule
\end{tabular}
}
\caption{Comparison of single-run performance with different MoE balance coefficients $\lambda$ on R2R, REVERIE, and \objnav-MP3D.}
\label{tab:balance}
\end{table}
We investigate the effect of MoE routing balance loss coefficient $\lambda$ in Table \ref{tab:balance}. When $\lambda$ is large, the balancing loss would force the model to evenly select different experts for samples in the same batch, when $\lambda$ decreases, such constraint is weakened. The model trained with $\lambda=0.2$ in row 1 performs worse than all other variants, demonstrating the significance of the balancing loss in MoE training. 
\section{Related Work}

\subsection{Vision-and-Language Navigation}
The development of a navigation agent capable of interpreting and acting upon unrestricted linguistic instructions to navigate through unfamiliar, photorealistic environments has been a longstanding goal within the field of Vision-and-Language Navigation~\cite{anderson2018vision, anderson2020rxr, qi2020reverie, thomason2020cvdn, zhu2021soon, zhang2024vision, hong2023learning}. Approaches to this problem have primarily addressed two main areas:
(1) Vision-Language Alignment: Some studies~\cite{hao2020prevalent, hong2020recurrent, guhur2021airbert, majumdar2020improving, chen2021hamt, zhao2023mind, qiao2023hop+, li2019robust, li2021improving} leverage generic visual-linguistic representations~\cite{chen2020uniter, li2020oscar, su2019vl, li2019visualbert, tan2019lxmert}. Others incorporate additional supervision through data augmentation~\cite{anderson2018spl, tan2019envdrop, li2022envedit, parvaneh2020counterfactual, wang2023scaling, guhur2021airbert, dou2022foam, li2023vlnsig, li2023panogen}, along with training strategies~\cite{wang2018look, wang2019reinforced, ma2019self, zhu2020vision, huang2019transferable} to enhance cross-modal alignment. Recently, some work~\cite{pan2023langnav, zhou2024navgpt, chen2024mapgpt, long2023discuss, lin2024navcot, long2024instructnav, zhan2024mc, zhang2024navid, zheng2023towards, zhou2025navgpt2, qiao2023march, qiao2025copilot} integrate LLMs with navigation policy for generalizable world knowledge. (2) Efficient Action Planning Mechanisms: These involve historical state memorization~\cite{hong2020recurrent, chen2021history}, self-correction strategies~\cite{ke2019tactical, ma2019regretful}, navigation map construction~\cite{an2021neighbor, chen2021topological, chen2022duet, wang2020active, zhao2022target, liu2023bird, wang2023gridmm, liu2024volumetric}, and the use of external knowledge prompts~\cite{lin2022adapt, li2023kerm}.

\subsection{ObjectGoal Navigation}
Approaches to learning to understand visual semantics and perform object goal navigation~\cite{xia2018gibson, chang2017matterport3d, ramakrishnan2021hm3d, khanna2024hssd, kolve2017ai2, deitke2022prothor, deitke2020robothor, ehsani2021manipulathor, yenamandra2023homerobot, khanna2024goat, gadre2022cows} could be categorized into two streams: (1) Modular Pipelines with Learned Modules~\cite{chaplot2020neural, chaplot2020object, hahn2021no, ramrakhya2022habitatweb, ramakrishnan2022poni, wasserman2023last}: 
This paradigm integrates learning into specific modules by leveraging explicit scene representations like semantic map~\cite{ramakrishnan2022poni, chaplot2020neural, hahn2021no}, or simply employ object detectors or segmentors~\cite{ramrakhya2022habitatweb, maksymets2021thda}. (2) End-to-end Learning with RL or IL~\cite{al2022zero, ye2021auxiliary, ramrakhya2023pirlnav, wijmans2020ddppo, yadav2023offline}: these methods benefit from visual representation~\cite{mousavian2019visual, yang2018visual}, auxiliary task~\cite{ye2021auxiliary}, and data augmentation~\cite{maksymets2021thda} to generalize to unseen environments.

\subsection{Mixture of Experts}
Mixture of Experts (MoE)~\cite{jacobs1991adaptive,jordan1994hierarchical,shazeer2017outrageously} models utilize multiple specialized experts along with a routing network that dynamically assigns tasks based on their complexity. 
Task-oriented MoE enhances the model's capacity to learn both specific and shared knowledge in computer vision, without significantly increasing computational costs~\cite{riquelme2021scaling,mustafa2022multimodal,shen2023scaling,chen2023octavius,gou2023mixture,liu2024task}.
Sparsely activated MoE is widely employed in LLMs to reduce computational costs while enabling the training of gigantic models with trillions of parameters through sparse activation~\cite{shazeer2017outrageously,lepikhin2020gshard,fedus2022switch}. MoE assigns different experts for instruction tuning in recent LLMs~\cite{xue2024openmoe,jiang2024mixtral,dai2024deepseekmoe}, optimizing the learning process by focusing on specific tasks within the overall model architecture.
To address task conflicts and enhance generalization in unseen tasks, some methods~\cite{wang2022adamix,chen2023octavius,gou2023mixture} employ various strategies to optimize the selection and aggregation of expert outputs in MoE.
\section{Conclusion}

This paper unifies a diverse range of navigation tasks within a cohesive language-guided navigation framework. It examines the fundamental challenges of sharing common knowledge while leveraging task-specific capabilities in navigation learning. We propose the State-Adaptive Mixture of Experts (SAME), which enables an agent to make decisions by integrating multi-granularity language inputs and dynamic observations. We believe SAME can guide the learning towards versatile language-guided navigation agents.
\section*{Acknowledgements}
This work was supported by the Centre for Augmented Reasoning, an initiative by the Department of Education, Australian Government.

{
    \small
    \bibliographystyle{ieeenat_fullname}
    \bibliography{main}
}

\clearpage
\maketitlesupplementary

\appendix

In this supplementary material, we aim to provide additional details to support the main content of our paper:

\begin{itemize}
    \item \textbf{Comparison of Training Methods:} In Section \ref{sec:comparison}, we compare and discuss the training methodologies employed in \textit{ObjectGoal Navigation (\objnav)} and \textit{Vision-and-Language Navigation (VLN)} research. This comparison highlights the motivation behind the chosen training strategies and the data selection for \ours.
    \item \textbf{Illustration of the DUET Method:} Section \ref{sec:duet} offers a comprehensive explanation of the DUET~\cite{chen2022duet} approach utilized in our study, elucidating its design and integration within our framework.
    \item \textbf{Full Results:} The complete results of \ours\ are provided in Table \ref{tab:full}, showcasing the performance of our method across various metrics and datasets.
\end{itemize}

\section{\objnav and VLN Training}
\label{sec:comparison}

Besides learning shareable knowledge and task-specific skills from the model design of \ours, another challenge under the unified language-guided navigation framework is to determine the most effective approach to facilitate the learning of agents’ language comprehension capacity and grounding it in action prediction. Analyzing the rationality within the contrasting research focuses on ObjectGoal Navigation and VLN offers insights into this challenge. Specifically, we observe that the primary differences lie in the \textit{\textbf{training data}} and \textit{\textbf{training methods}} used. In this section, we discuss and make strategic decisions in \ours\ training regarding these two aspects, to address the above challenge.

When language instructions are minimal, the task reduces to \objnav~\cite{batra2020objectnav}, where the learning objective is the semantic affinity between the target object and the visual perception, and leveraging episodic memory for strategic exploration without redundant revisits, since no extract information is provided from the language instruction. From the \textit{\textbf{data}} aspect, it is proven to be effective to learn 
strategical exploration through human demonstration, and data collection is done in several works~\cite{ramrakhya2022habitatweb, ramrakhya2023pirlnav}. From the \textbf{\textit{training}} aspect, \objnav combines learning “where” and “how” to move, incorporating semantic perception and low-level collision avoidance (\texttt{FORWARD}, \texttt{TURN\_LEFT}, \texttt{TURN\_RIGHT}, \texttt{STOP}) within a continuous environment~\cite{savva2019habitat}.

On the contrary, VLN requires higher-level language understanding, where the agents not only need to understand the visual semantics of the environment but also need to align past observation and action sequence with the language description. 
From the \textit{\textbf{data}} aspect, VLN agents learn higher-level language comprehension capacity from human-annotated instructions for navigation episodes. 
From the \textit{\textbf{training}} aspect, such alignment is hard to learn directly in a continuous environment, evident by the low performance ($\sim 35 \%$ SR) on VLN-CE benchmark~\cite{krantz2020beyond} of the methods that directly operate in continuous environments. Therefore, VLN research typically decouples Vision-Language alignment from collision avoidance by learning to navigate in a discretized environment~\cite{anderson2018r2r}, where the navigable viewpoints are densely sampled from the environment at an average separation of 2.25m to form a navigation graph. The learned multimodal navigation policy performs high-level action by selecting view directions that contain a navigable viewpoint and teleporting between viewpoints on the graph. A waypoint predictor~\cite{krantz2021waypoint, krantz2022sim, hong2022bridging} is employed to bridge action space discrepancies in continuous settings. Decoupling the learning of Vision-Language-Action alignment with low-level action control significantly benefits the learning of language understanding capacity, improving VLN-CE success rates by $\sim$ 20\%.

To bridge action space discrepancies in continuous settings, modular designs employ waypoint predictors to propose navigable waypoints based on current observation, while the multimodality navigation policy performs view selection conditioned on these waypoints, with a heuristic controller executing low-level actions to move to the waypoint. Decoupling the learning of vision-language-action alignment with low-level action control significantly benefits the language understanding capacity, improving VLN-CE success rates by approximately 20\%. In this work, we hypothesize such modular setups optimize the learning of language understanding capacity, which guides us to perform unified policy training in the discrete environment.

Building on the aforementioned discussion, this work concentrates on solving the high-level decision-learning problem by decoupling it from tasks such as collision avoidance and low-level control. This direction motivates the adoption of \textit{\textbf{VLN training methods}} for \ours\ training within a discrete environment. 
Regarding \textit{\textbf{training data}}, we combine \objnav human demonstration data with VLN human-annotated instructions to capture and learn distinct navigation behaviors.

\section{DUET Revisit}
\label{sec:duet}

\ours\ builds upon the design of the Dual-scale Graph Transformer (DUET)~\cite{chen2022duet}. DUET incorporates a text encoder to process instructions and employs both global and local branches to facilitate cross-modal reasoning at coarse and fine scales.

\subsection{Text and Visual Embedding}

DUET’s text encoder leverages a 12-layer transformer, initialized with LXMERT~\cite{tan2019lxmert}. For visual embedding, each node’s visual observation comprises 36 view images, covering 12 horizontal and 3 vertical directions. To differentiate between these views, a directional embedding $E^{ang}$ is added to the visual features $\hat{\mathcal{O}_t}$, which are extracted by the vision encoder. Since DUET incorporates all 36 view images to construct the spatial observation, navigable adjacent nodes are only visible in a subset of these views, referred to as navigable views. To account for this, a navigable embedding $E^{nav}$ is also included. The final visual embedding is processed by a 2-layer transformer to encode spatial relationships between views, producing panoramic view embeddings:

\begin{equation}
\mathcal{O}^\text{pano} = \text{SelfAttn}\left( \hat{\mathcal{O}_t} + E^\text{ang} + E^\text{nav}\right).
\end{equation}

\subsection{DUET Local Branch}

This section focuses on the local branch of DUET, which predicts actions based on the current node’s instruction and egocentric observation. Unlike the global branch, no graph-level information is utilized beyond local observations.

\subsubsection{Local Visual Embedding}
The panoramic view embedding $\mathcal{O}^\text{pano}$ is augmented with two types of location embeddings. The first represents the relative location of the current node with respect to the starting node, encoding long-distance directional relationships. The second represents the egocentric directions of adjacent views at the current node, enabling actions such as “turn right.”

\subsubsection{Local Cross-Modal Encoding}

The local branch employs a standard 4-layer cross-modal transformer to capture relationships between vision and language. During action prediction, a mask is applied to exclude unnavigable views and action logits are computed only for the navigable views at the current node.

\subsection{DUET Global Branch}
This section introduces the global branch of DUET, which tasks the topological map representation $\hat{\mathcal{G}_t}$ and encoded language instruction $\hat{\mathcal{W}}$ to predict actions by selecting any nodes on the graph.

\subsubsection{Node Embedding}
For each node on the graph, two additional encodings are applied: a location encoding $E^\text{loc}$ and a navigation step encoding $E^\text{step}$. The location encoding represents the egocentric position of a node on the map, capturing its orientation and distance relative to the current node. On the other hand, the navigation step encoding assigns a value corresponding to the latest visited timestep for previously visited nodes, while unexplored nodes are encoded with a value of 0. This encoding scheme enables the model to differentiate nodes based on their navigation history, thereby enhancing alignment with the provided instructions. Additionally, a special “stop” node is introduced into the graph to signify the stop action. This node is connected to all other nodes in the graph.

\subsubsection{Global Cross-Modal Encoding}
The encoded node features and word embeddings are processed through a 4-layer graph-aware cross-modal transformer, which is composed of the following two key components, as illustrated in Figure \ref{fig:pipeline}.

\noindent\textbf{Cross-Attention Layer} This layer models the relationships between the global map and the instruction, enabling cross-modal alignment. \ours\ examine applying State-Adaptive MoE on the visual query $W_q$ or textual key $W_k$ and value $W_v$ in this layer.

\noindent\textbf{Graph-Aware Self-Attention Layer (GASA)} Unlike standard self-attention mechanisms which rely solely on visual similarity, the GASA module incorporates the graph’s structural information to refine attention computation, formulated as follows:
\begin{equation}
    \text{GASA}(\mathcal{V}) = \text{Softmax}\left( \frac{\mathcal{V}W_q(\mathcal{V}W_k)^T}{\sqrt{d}}+A(\mathcal{E}_t)\right)\mathcal{V}W_v,
\end{equation}
where $A(\mathcal{E}_t)$ represents the spatial affinity matrix, comprised of pairwise L2 distances among all observed nodes.
By incorporating this spatial context, GASA ensures that the model prioritizes spatially or topologically proximate nodes, which are often more contextually relevant than visually similar but distant nodes.

Each block in the global branch concludes with a Feed-Forward Network (FFN). Additionally, \ours\ explores applying the State-Adaptive MoE mechanism to this FFN, as depicted in Figure \ref{fig:pipeline} of the main paper.

\section{Full Results on All VLN Tasks}
We show the full results of \ours\ on all the tested VLN benchmarks in Table \ref{tab:full}.
\begin{table}[t]
\centering

\resizebox{\columnwidth}{!}{
\definecolor{Gray}{gray}{0.94}

\begin{tabular}{lccc>{\columncolor{Gray}}c>{\columncolor{Gray}}cccc>{\columncolor{Gray}}c>{\columncolor{Gray}}c}
\toprule
\midrule
\multicolumn{1}{c}{\multirow{2}{*}{Benchmark}} & 
\multicolumn{5}{c}{Val Unseen} & 
\multicolumn{5}{c}{Test Unseen} \\ 

\cmidrule(r){2-6}
\cmidrule(r){7-11}

\multicolumn{1}{c}{} & 
\multicolumn{1}{c}{TL} & 
\multicolumn{1}{c}{NE$\downarrow$} & 
\multicolumn{1}{c}{nDTW$\uparrow$} & 
\multicolumn{1}{c}{SR$\uparrow$} & 
\multicolumn{1}{c}{SPL$\uparrow$} & 
\multicolumn{1}{c}{TL} & 
\multicolumn{1}{c}{NE$\downarrow$} & 
\multicolumn{1}{c}{GP$\uparrow$} &
\multicolumn{1}{c}{SR$\uparrow$} & 
\multicolumn{1}{c}{SPL$\uparrow$} \\

\midrule
\midrule

R2R~\cite{anderson2018r2r}
& 13.65 & 2.73 & 71.05 & 76.25 & 66.16
& 14.80 & 3.03 & -- & 73.92 & 64.41 \\
RxR-EN~\cite{anderson2020rxr}
& 22.69 & 6.53 & 51.20 & 50.52 & 42.19
& -- & -- & -- & -- & -- \\
REVERIE~\cite{qi2020reverie}
& 18.87 & 5.18 & 48.54 & 46.35 & 36.12
& 19.47 & -- & -- & 48.60 & 37.10 \\
SOON~\cite{zhu2021soon}
& 34.42 & 8.12 & -- & 36.11 & 25.42
& 37.99 & -- & -- & 38.18 & 27.11 \\
CVDN~\cite{thomason2020cvdn}
& 30.90 & 12.72 & -- & 24.48 & 17.23
& -- & -- & 7.07 & 18.15 & 12.18 \\

\bottomrule
\end{tabular}
}
\caption{Full results of \ours\ on all VLN benchmarks.}
\label{tab:full}
\end{table}

\end{document}